\documentclass{article}
\usepackage{spconf,amsmath,graphicx}
\usepackage{subcaption} 
\usepackage{amsfonts}       
\usepackage{breqn}

\newcommand{\norm}[1]{\left\lVert#1\right\rVert}

\title{Crime Event Embedding with\\Unsupervised Feature Selection}
\name{Shixiang Zhu, Yao Xie}
\address{Georgia Institute of Technology}
\begin{document}
\maketitle
\begin{abstract}
We present a novel event embedding algorithm for crime data that can jointly capture time, location, and the complex free-text component of each event. The embedding is achieved by regularized Restricted Boltzmann Machines (RBMs), and we introduce a new way to regularize by imposing a $\ell_1$ penalty on the conditional distributions of the observed variables of RBMs. This choice of regularization performs feature selection and it also leads to efficient computation since the gradient can be computed in a closed form. The feature selection forces embedding to be based on the most important keywords, which captures the common modus operandi (M. O.) in crime series. Using numerical experiments on a large-scale crime dataset, we show that our regularized RBMs can achieve better event embedding and the selected features are highly interpretable from human understanding. 
\end{abstract}
\begin{keywords}
RBM, $\ell_1$ regularization, feature selection
\end{keywords}
\section{Introduction}
\label{sec:intro}

Event data such as crime incidents, medical reports, and social media data \cite{triovecevent.zhang, event2vec.hong} is increasing rapidly in the era of big data. Each event data sample usually consists of time, location, and other description of the event such as free text. Learning correlation between events has become a pressing need for event data. Embedding is an efficient solution to represent unstructured events data by mapping the discrete events into a continuous Euclidean space, and the distance between two data points is proportional to their similarity or correlation \cite{word2vec.mikolov}. Recently, research on embedding has received a lot of attention. The central problem is to learn the mapping, which has been achieved via various approaches including neural networks such as Restricted Boltzmann Machine. 

However, a major challenge remains in how to handle free text of the event. For crime data, the free-text part may actually contain the most important information. Current embedding approaches based on tokenizing text part as simple "marks" of events \cite{Du2016} may not be sufficient to capture these useful information since they are buried in the complex and unstructured free-text narratives of the police event report. To achieve the goal, we need to develop models and tools that combines embedding and natural language processing. 

In this paper, we present a novel event embedding approach that jointly captures the free-text part of the event data together with other structured data (such as time and location). We focus on crime event data, where each event is a police report that contains time, location, and narrative of the crime incident entered by the investigator.  We leverage the Restricted Boltzmann Machines' (RBMs) structure to perform embedding and introduces a novel way to regularize RBM to automatically select the most important observed (visible) variables in order to improve the quality of the embeddings. 

The motivation for the new regularization for variable selection in our work is two-fold. First, as shown in Appendix \ref{appen:motivation}\footnote{See https://arxiv.org/abs/1806.06095 for the appendix.}, it is motivated because we observe in crime data, very few of certain keywords in the free-text (which corresponds to the observed variables in RBMs) are the most important in defining the correlation between events. Thus, it is crucial to identify these keywords while eliminating the irrelevant ones. This can be viewed as performing "variable selection'' for RBM (as analogous to variable selection using lasso for linear regression model). We found that by focusing on the most important observed variables (implicitly through regularization), the performance of embedding will be drastically improved. Second, this regularizes the over-parameterization of RBM (as typically there is a large number of observed and hidden variables) in the presence of limited data. Since we do not know which keywords are the most important {\it a priori} in determining the correlation of events,  we extracted as much as possible from the free-text and this may lead to over-representation. To a certain extent, the regularization automatically get rid of redundant information and only select the most crucial variables (keywords). 

Our method is {\it unsupervised}: training RBM does not require any "label" of events since RBM essentially captures the co-occurrence pattern of the observed variables. Thus, our work can be viewed as an unsupervised variable selection for embedding purposes. This can be quite important because, in crime data analysis, it is hard for police officers to provide a large number of crime incidences that are known to be related to each other. In our later example, we are only able to identify 6 groups of labeled data for verification purposes only but not for training purposes. To the best of our knowledge, our work is the first to demonstrate the value of embedding based on RBM for finding correlated cases in crime data. Although in the paper, we focus on crime event analysis, this framework can be generalized to other event embeddings as well. 

\section{Related Work}
\label{sec:related}

Most existing events embedding methods focus on modeling spatial-temporal dependencies between events \cite{event2vec.hong} for the location and time information of the events, by building (hierarchical) probabilistic models. Some recent work also takes into account the actual content of the events, for example, the type (or category) of events \cite{Du2016}. This is done by using Recurrent Neural Networks (RNN) treating category as marks of events, and thus far usually only a small number of possible (discrete) marks are considered in the model. However, this can be restrictive for some real data, as it cannot handle the case when there is a much richer textual description for each event, and it is not easy to extend existing work based on ``marks'' to handle rich textual event data.   

To perform event embedding in the presence of textual data, a mainstream approach is the so-called Skip-gram method \cite{skipgram.mikolov, item2vec.barkan, factorembedding.liang, triovecevent.zhang}. Skip-gram was originally introduced to model the semantic structure of the language  \cite{word2vec.mikolov}.  Generally, it can capture the co-occurrence patterns of words in each document. However there is no co-occurrence concept in crime analysis since the manpower of police is too limited and expensive to define enough context in order to calculate the co-occurrence of the crime events. 

By and large, this is an extensive work based on our paper \cite{crime.zhu}. In addition to previous work, this paper extensively considers the regularization for RBM in order to select key features from the observed variables. There is also some other good work take into account regularization for RBM for different purposes. The close work \cite{Luo2010, Halkias2013} propose various of regularization on the hidden variables of RBM for yielding sparsity in hidden variables. Another work \cite{Ranzato2007, Ranzato2008} came up with the similar idea of sparse feature learning by imposing regularization on RBM structure. However, the proposed deep architecture is different from our method and it mainly focuses on producing good representations with deep feature hierarchies, which is completely different from the motivation of feature selection.

\section{Problem Setup}
\label{sec:pagestyle}

A single event data point consists of a set of observed variables $\mathcal{X} = \{x_s, x_t,\} \cup \{x_i\}_{i\in\mathbb{Z}^{+}, i \le V}$, where $x_s \in \mathbb{R}^{D}$ ($D$ is the dimension of the space), $x_t \in \mathbb{R}$ are temporal and spatial variables respectively, which are explicitly retained in the model (meaning that they will not be eliminated by the variable selection). And $\{x_i\}_{i\in\mathbb{Z}^{+}, i \le V}$ is a set of observed variables represent the tf-idf value \cite{nlp.stanford} of the keywords in the vocabulary that appeared at least once in the corpus where $x_i \in \mathbb{R}$, $i$ indicate the index of the keyword and $V$ is the total number of the keywords. In our model, $\mathbf{x} = \{x_i\}_{i\in\mathbb{Z}^{+}, i \le V}$ will be regularized in order to select key variables. Each event in the dataset is denoted as a triplet $(x_s, x_t, \mathbf{x})^{(k)}$ where $k$ is the index of the event. 

Given an event point $(x_s, x_t, \mathbf{x})$, we define its embedding as $\mathbf{h} \in \{0,1\}^H$ with the dimension of $H$ (in our later examples we set $H = 1,000$). The similarities between two embedding vector can be evaluated by their cosine distance $\mathbf{h}\cdot \mathbf{h}\textprime  / \norm{\mathbf{h}} \cdot \norm{\mathbf{h}\textprime}$, where $\|\mathbf h\|$ denotes the $\ell_2$ norm of vector $\mathbf h$. The goal  is to take the event dataset $\{(x_s, x_t, \mathbf{x})^{(k)}\}$ as input and produce their embeddings $\{\mathbf{h}^{(k)}\}$ accordingly. 

\section{Regularized RBM}

In this section, we present our new regularized RBM with feature selection. A basic introduction to the vanilla RBM and its related notations are delegated to the Appendix \ref{appen:rbm}\footnote{See https://arxiv.org/abs/1806.06095 for the appendix.}. 

\subsection{Observed variables selection in RBM}

We introduce a $\ell_1$-regularizer to the log-likelihood of RBM (\ref{eq:log_likelihood}) to mitigate the impact of those noisy variables specifically. As discussed in Section 1, directly learning the statistical dependencies between all observed variables (the Bag-of-Words in the corpus) will bring noisy information from irrelevant variables into the model. To achieve the selection, we impose an $\ell_1$ penalty on the probability $1 - P(x_i < t | \mathbf{x})$\footnote{Here $t$ is a very small constant, we preset $t=10^{-2}$ here since the impact of a tf-idf value lower than $10^{-2}$ can usually be ignored.} weighted by $\lambda$, which penalizes the reconstructed observed variables that are sensitive to large values. This penalty introduces a natural way to select the most important features (correspond to observed variables in RBM). 
A nice feature of this penalty is that the corresponding gradient can be computed easily. 

Thus, given one training data $\mathbf{x}$, we need to solve the following optimization problem. {\it This leads to {our new formulation which performs the selection of observed variables for RBM}:}
\begin{equation}
\begin{aligned}
\max_{\mathbf{w}, \mathbf{b}, \mathbf{c}} \Bigl \{ 
&\log\ \mathcal{L}(\theta|(x_s, x_t, \mathbf{x})) - \\
&\lambda \sum_{i \le V} \left | 1- P(x_i < t | \mathbf{x})\right | \Bigr \}
\end{aligned}
\label{eq:new_obj}
\end{equation}
By tuning the weight $\lambda$ of the penalty, we can achieve various levels of sparsity in terms of a subset of observed variables $\{x_i\}, i \le V$. 

We solve this optimization problem by gradient descent (note that this is a non-convex problem and gradient descent is a default approach to solve it). Because, in crime analysis, the observed variables are real values (tf-idf), we take the Gaussian-Bernoulli RBM (GBRBM) as an example. $1 - P(x_i < t | \mathbf{x})$ can be rewritten into the following expression by substituting Eq.(\ref{eq:condp.x}) (other types of RBM can be processed in a similar fashion):
\begin{align*}
& \left | 1 - P(x_i < t | \mathbf{x}) \right |= 1 - \\ 
& \int_{-\infty}^{t}  \frac{1}{\sigma \sqrt{2 \pi}} \cdot
e^{-\frac{1}{2 \sigma^2} (x_i - b_i - \sigma \sum_{j=1}^{n} h_j w_{ij})^2}  dx_i.
\end{align*}
We can directly compute its gradients with respect to $w_{ij}$ and $b_{i}$ in a closed-form due to the exponential structure in the second term of it log-likelihood function. By introducing this penalty term, the gradients $\nabla w_{ij}$ and $\nabla b_{i}$ in Eq.(\ref{eq:grad.w}) and Eq.(\ref{eq:grad.b}) can be rewritten as follows:
\begin{align*}
\nabla w_{ij} 
& =\left < x_i h_j\right >_{p(\mathbf{h}|\mathbf{x})} - \left < x_i h_j\right >_{p(\mathbf{x}, \mathbf{h})} \\ 
& - \lambda \frac{h_j}{\sqrt{2\pi}} \sum_{i \le V} \exp(-\frac{1}{2\sigma^2} (t- b_i - \sigma \sum_{j=1}^{H} h_j w_{ij})), \\
\nabla b_{i} 
&= x_i - \left < x_i \right >_{p(\mathbf{x})} \\
& - \lambda \frac{1}{\sigma \sqrt{2\pi}} \sum_{i \le V} \exp(-\frac{1}{2\sigma^2} (t-b_i-\sigma \sum_{j=1}^{H} h_j w_{ij})).
\end{align*}

\subsection{Training and interpretation}

\begin{figure}
\centering
\begin{subfigure}[h]{0.32\linewidth}
\includegraphics[width=\linewidth]{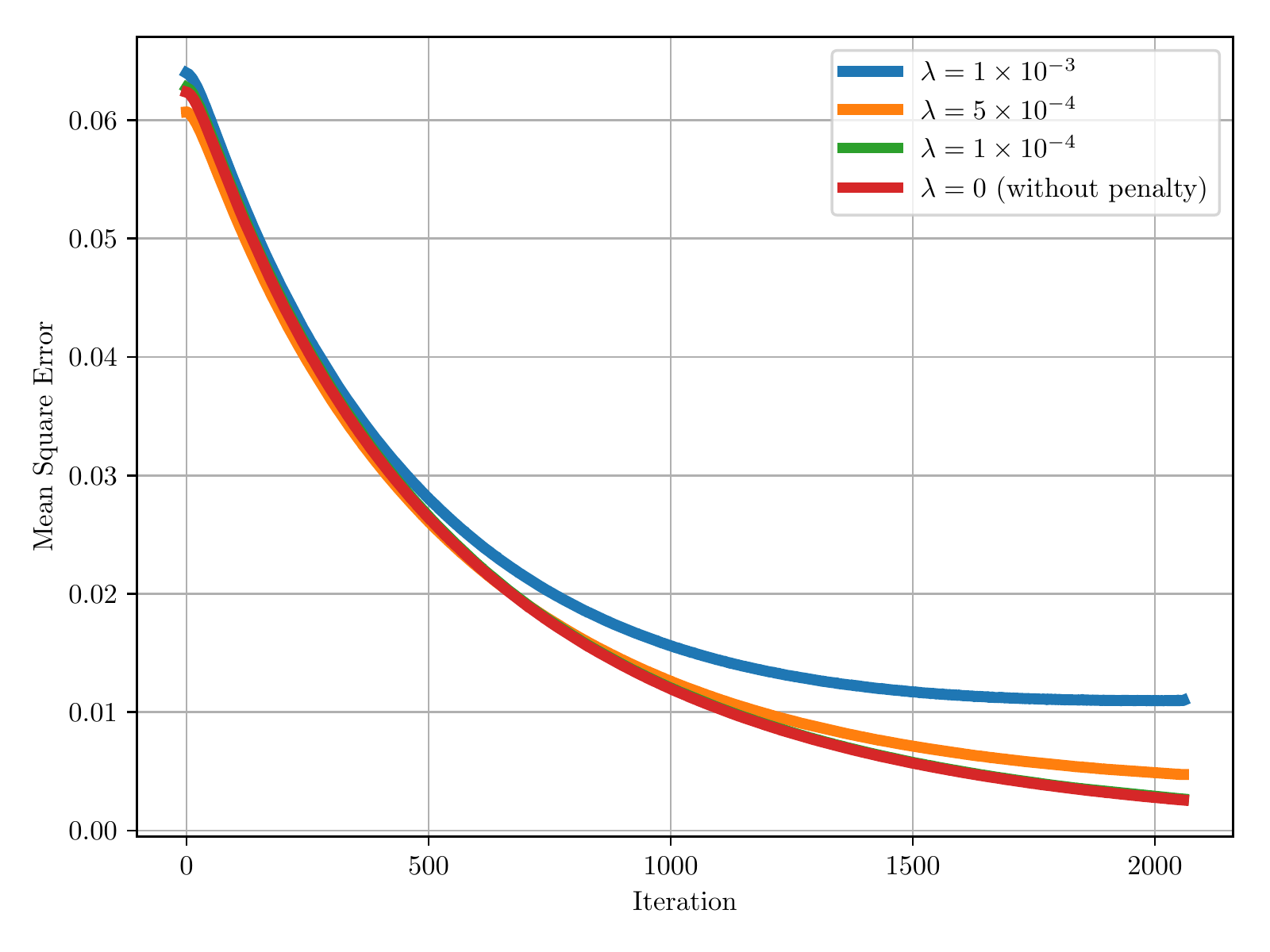}
\caption{}
\label{fig:error}
\end{subfigure}
\hfill
\begin{subfigure}[h]{0.32\linewidth}
\includegraphics[width=\linewidth]{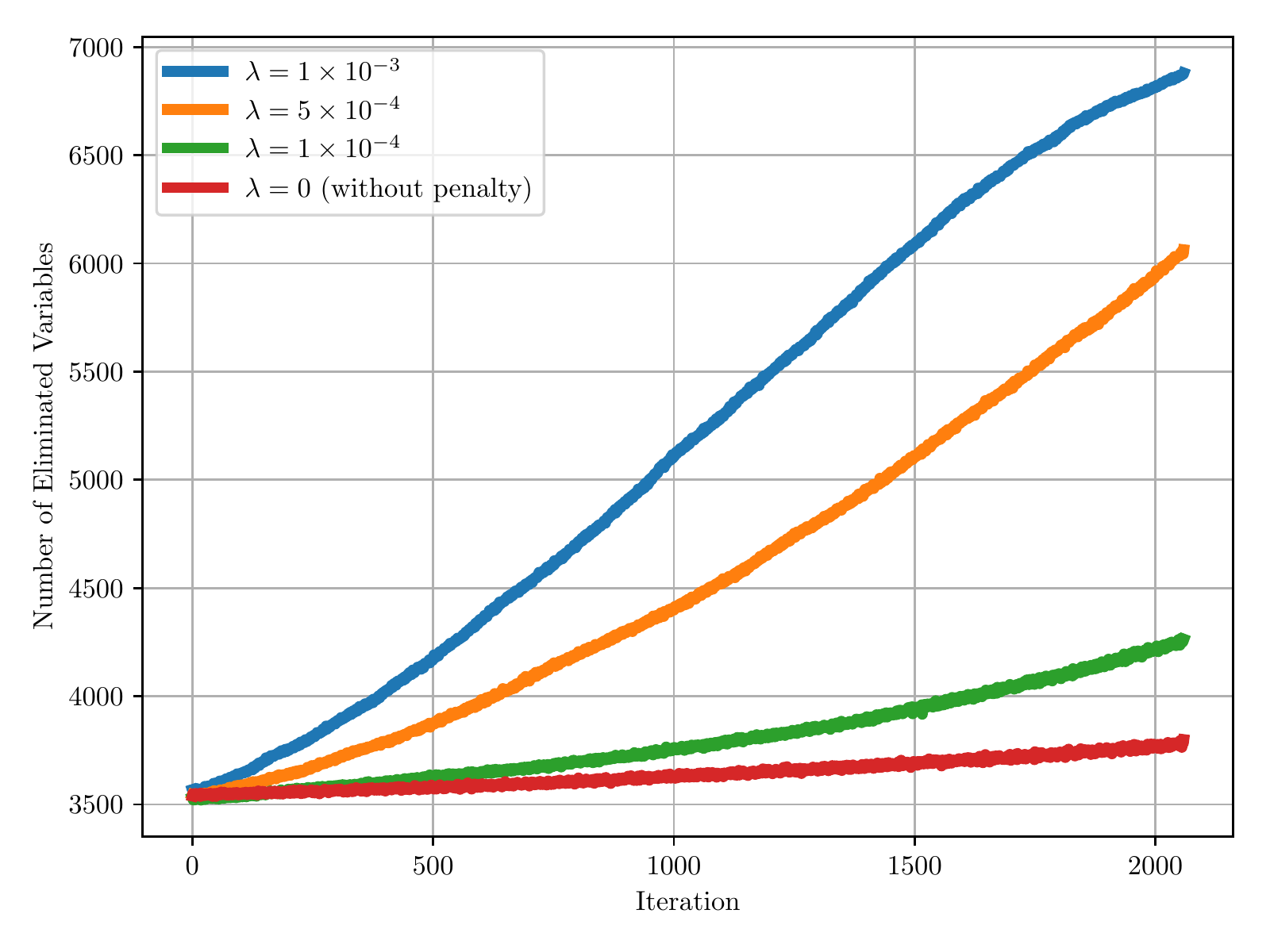}
\caption{}
\label{fig:zero}
\end{subfigure}
\hfill
\begin{subfigure}[h]{0.32\linewidth}
\includegraphics[width=\linewidth]{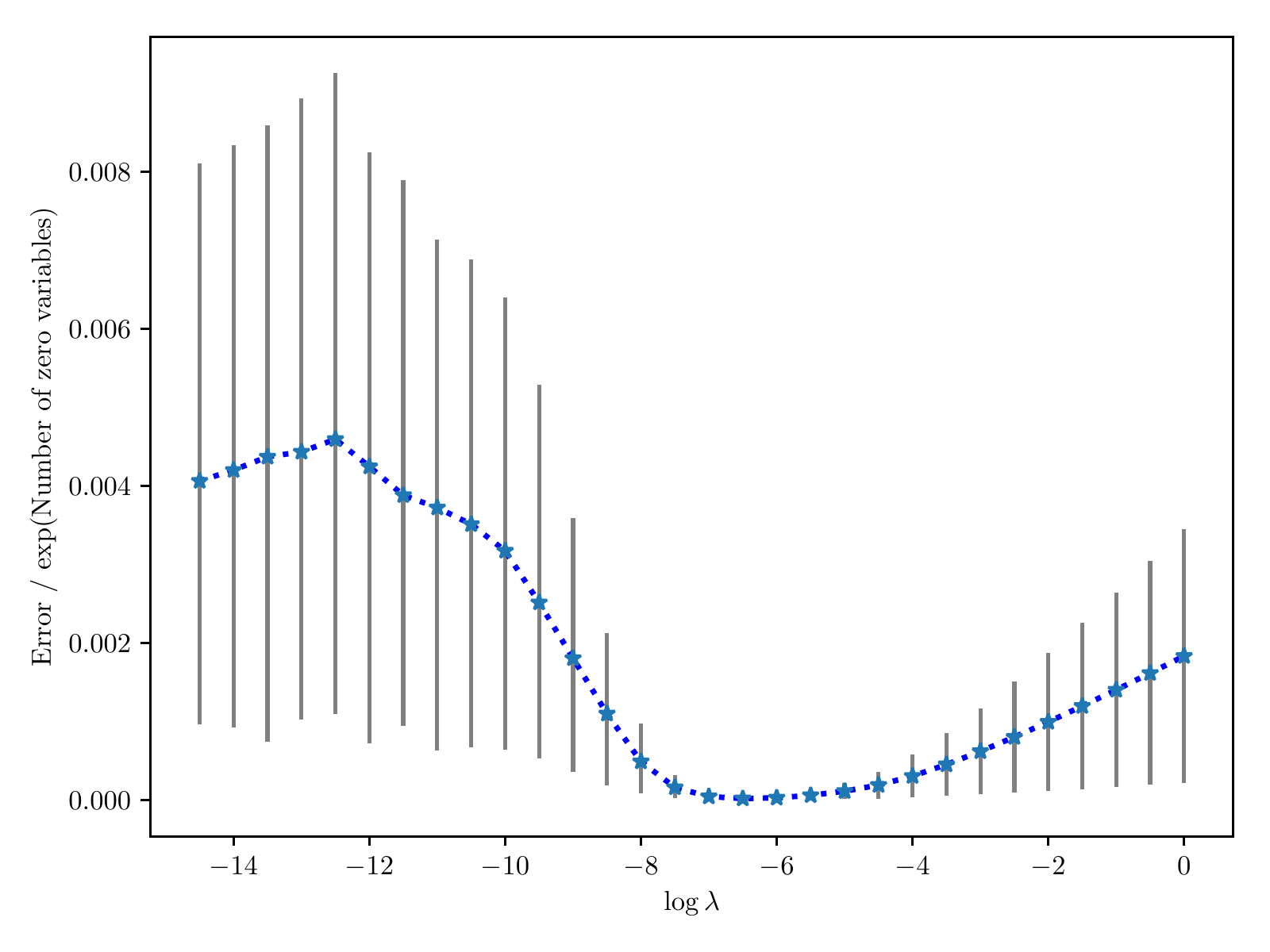}
\caption{}
\label{fig:cv}
\end{subfigure}
\caption{Fitting RBM with and without designed penalty term over 2,056 crime events (including 7,038 observed variables). Under same experiment settings (learning rate $\alpha = 10^{-3}$, threshold value $t=10^{-2}$): (\ref{fig:error}): training errors over iterations; (\ref{fig:zero}): numbers of eliminated (disabled) variables over iterations, and (\ref{fig:cv}): result of cross-validation over $\lambda$.}
\label{fig:fitting}
\vspace{-0.2in}
\end{figure}
The standard training approach for RBM is still applied in the regularized RBM thanks to the explicit closed-form of $w_{ij}$ and $b_i$. The training objective of the regularized RBM is to maximize a likelihood function defined via the energy function defined in (\ref{eq:new_obj}). we adopt the $k$-step contrastive divergence (CD-$k$) approach for the training, which is an approach to approximate the gradient in training RBM via gradient descent \cite{hinton2002}, where the gradients have been defined above. The Gibbs chain is initialized with a training example of the training set and yields the sample after $k$ steps. The iterations are repeated until certain empirical convergence has achieved.


Regarding feature selection, the effects of the $\ell_1$ regularization can be interpreted intuitively. The $\ell_1$ norm yields sparsity within the given subset of observed variables, which means that some variables' norms are set zero, or their activation possibilities tend to produce zero values. During the course of training, only very few of observed variables in the subset will win this competition, and their impact to the embeddings will get enhanced, though meanwhile, the effect of irrelevant variables will vanish eventually. Therefore, for our application, we impose the penalty on the subset of observed variables that associate the input of the Bag-of-Words vectors and successively eliminate keywords due to the regularization. As shown in Fig.\ref{fig:error} and Fig.\ref{fig:zero}, the regularized RBM prominently disable the most of keywords in the vocabulary (only 280 keywords are retained after convergence) when $\lambda=10^{-3}$ without losing too much accuracy on training errors. And in Fig.\ref{fig:cv}, we also show a surrogate cross-validation result over $\lambda$ in which the model achieves the best performance at some point. Differs from feature selection in a linear model (e.g., through lasso), here we have to perform a \textit{reconstruction} (similar to back propagation) step on the observed variables. The benefits of denoising observed variables will be presented in the embeddings due to the symmetrical structure of RBM. 

\section{Experiments on real-data}
\label{sec:typestyle}

To provide a comprehensive validation, we test our approach on crime events data provided by the Atlanta Police Department over 2016 and 2017 with carefully picked $\lambda=10^{-3}$ according to the result of cross-validation in Fig.\ref{fig:cv}. We show the superiority of our method over other competitors under the same parameter settings, including the vanilla \textbf{RBM} without regularization, Latent Dirichlet Allocation (\textbf{LDA}), Singular Value Decomposition (\textbf{SVD}), \textbf{denoising Autoencoder}. Specifically, the objective goal of these methods is to generate embeddings for each data point with the length of 1,000. We are ultimately looking for the distributed representations for each of the crime events, and the representations would tend to be closer in Euclidean distance if the crimes follow the same M.O. (committed by the same arrestee, suspect, or criminal gang).

\textbf{Dataset.} The dataset for the experiments contains 2,056 crime events in total happened in Atlanta over past two years, which consists of 56 hand-labeled events that belong to 5 individual crime series committed by different arrestees, and 2,000 randomly selected unknown events. The overall crime events involve 123 crime categories. Each crime event mainly includes time, location (latitude and longitude), crime category and free text part. The most important part is free text, all the text have been preprocessed into a Bag-of-Words, including 7,039 keywords and 2,056 documents. 

\subsection{Visualization via t-SNE} 

\begin{figure}[!h]
\centering
\begin{subfigure}[h]{0.33\linewidth}
\includegraphics[width=\linewidth]{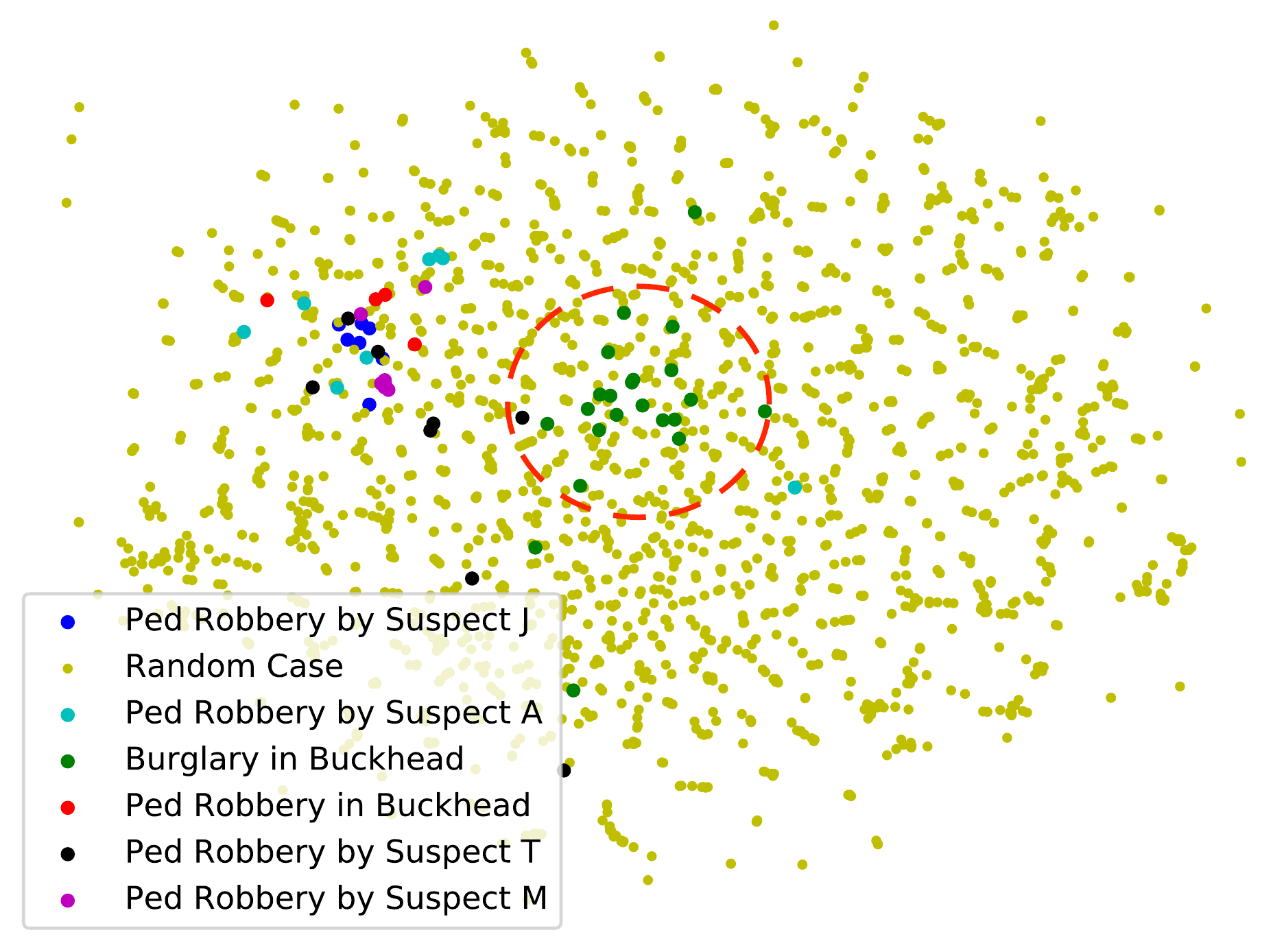}
\caption{regularized RBM}
\label{fig:lam1e-3}
\end{subfigure}
\hfill
\begin{subfigure}[h]{0.33\linewidth}
\includegraphics[width=\linewidth]{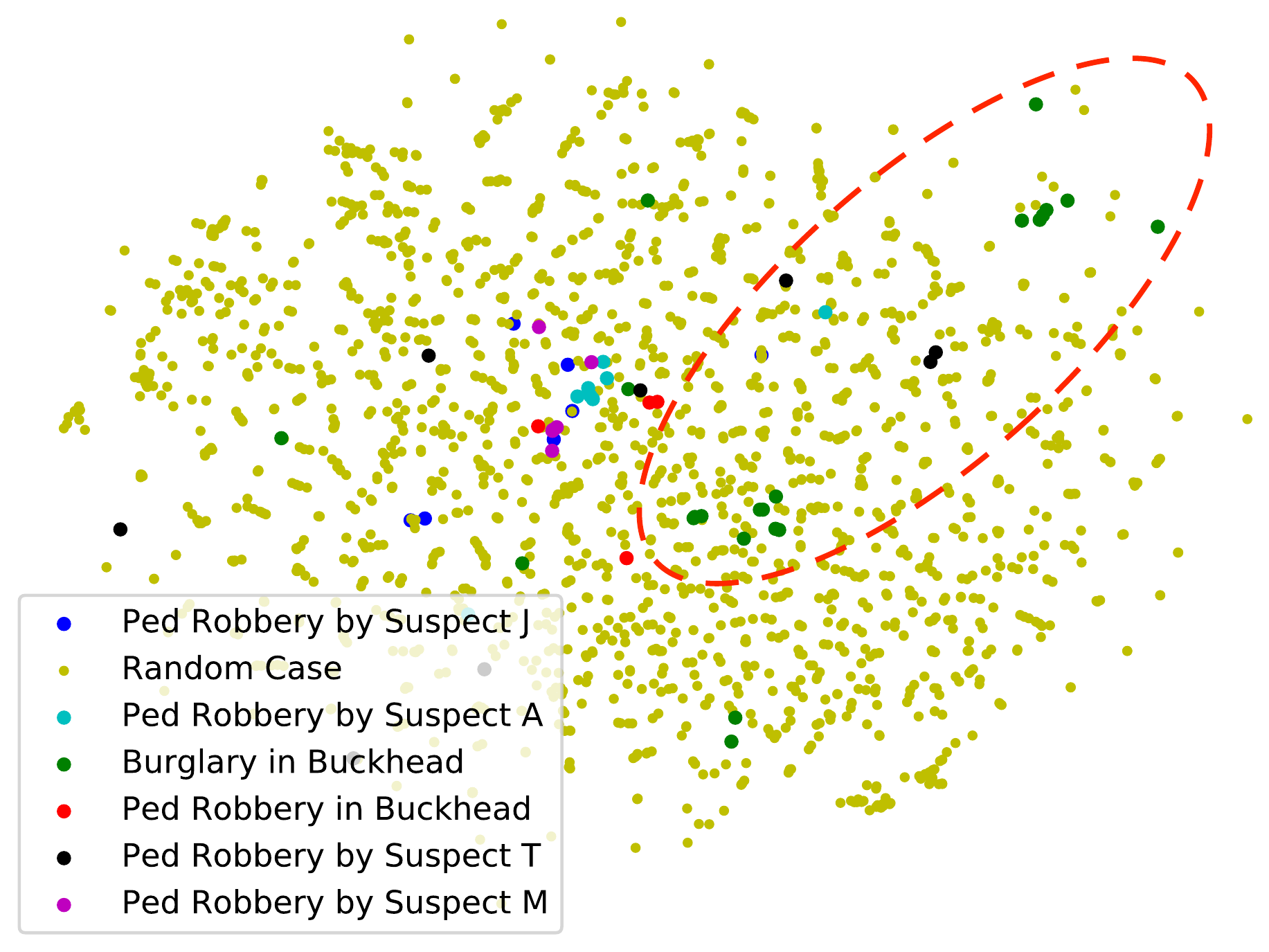}
\caption{RBM ($\lambda=0$)}
\label{fig:lam0}
\end{subfigure}
\hfill
\begin{subfigure}[h]{0.32\linewidth}
\includegraphics[width=\linewidth]{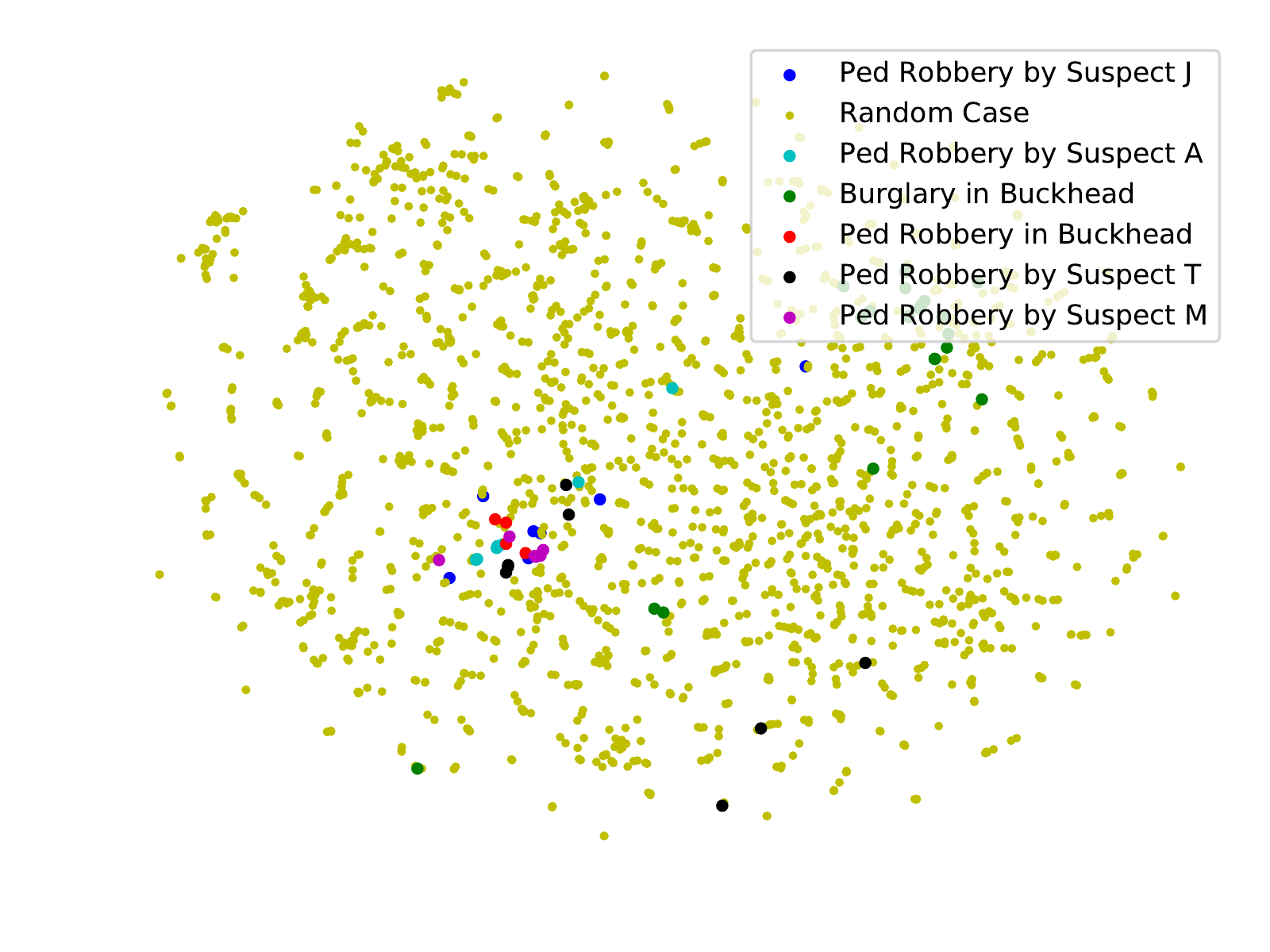}
\caption{SVD}
\label{fig:svd}
\end{subfigure}
\vfill
\begin{subfigure}[h]{0.32\linewidth}
\includegraphics[width=\linewidth]{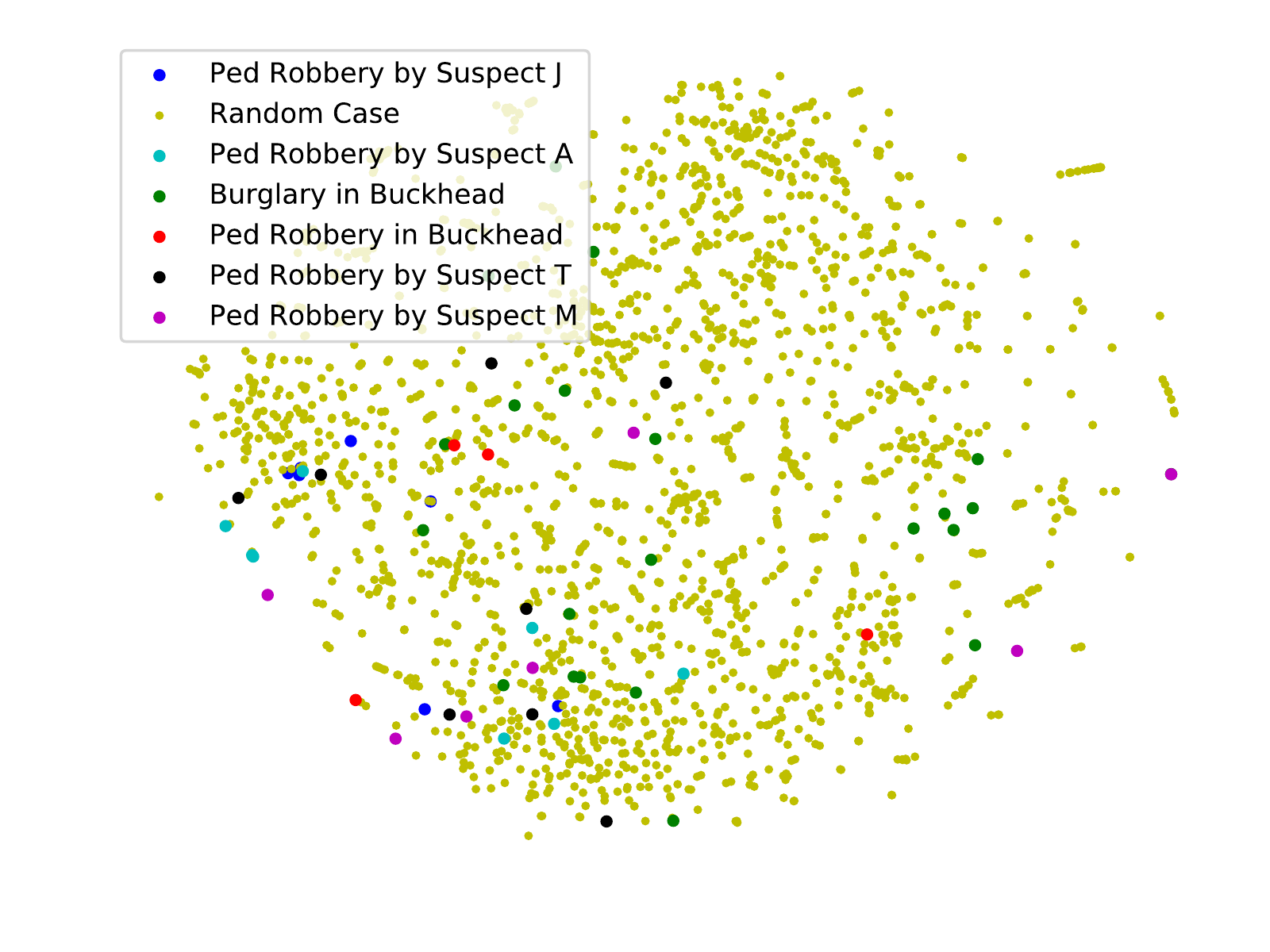}
\caption{LDA}
\label{fig:lda}
\end{subfigure}
\hfill
\begin{subfigure}[h]{0.32\linewidth}
\includegraphics[width=\linewidth]{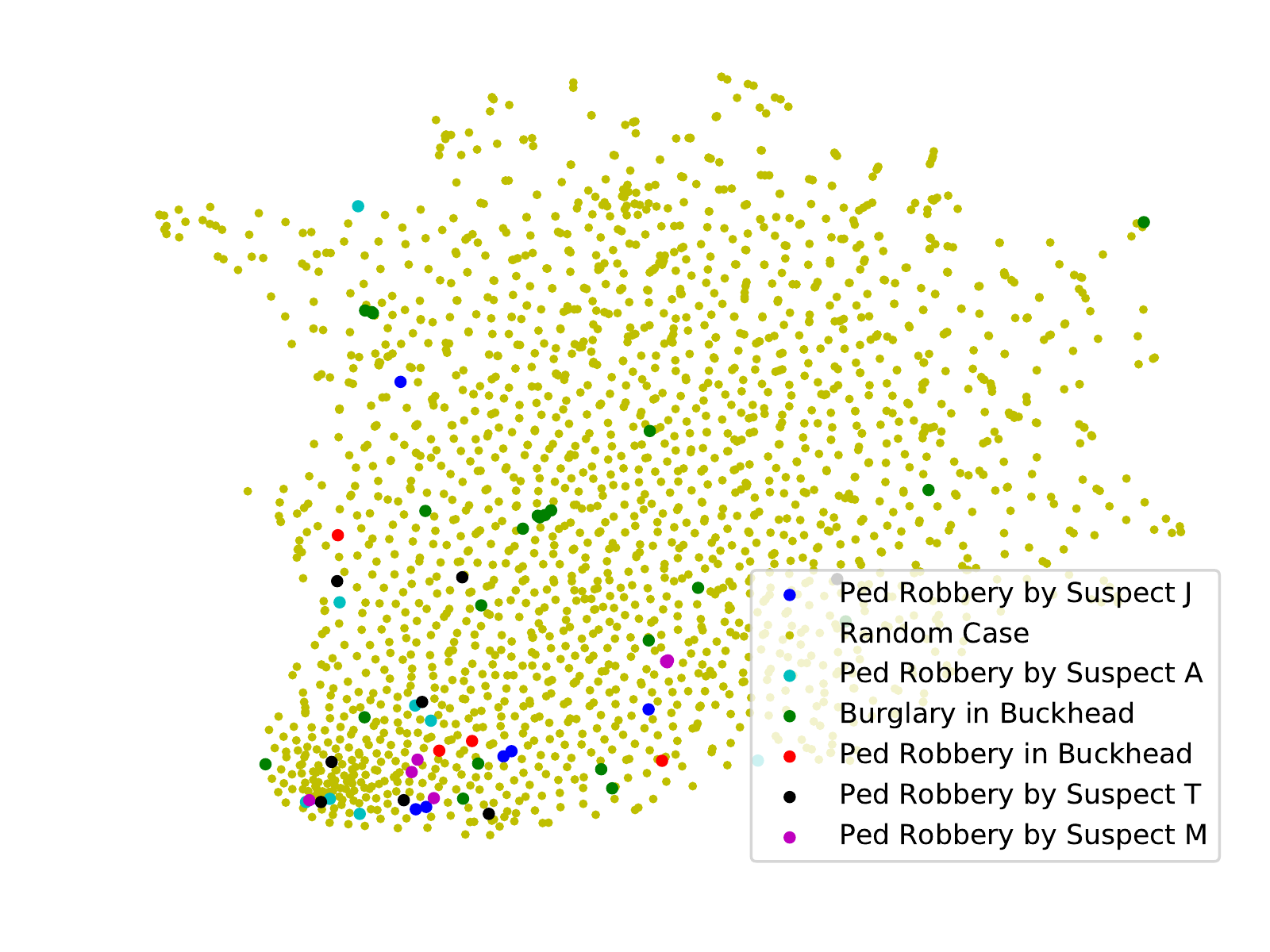}
\caption{Autoencoder}
\label{fig:autoencoder}
\end{subfigure}
\hfill
\begin{subfigure}[h]{0.32\linewidth}
\centering
\includegraphics[width=\linewidth]{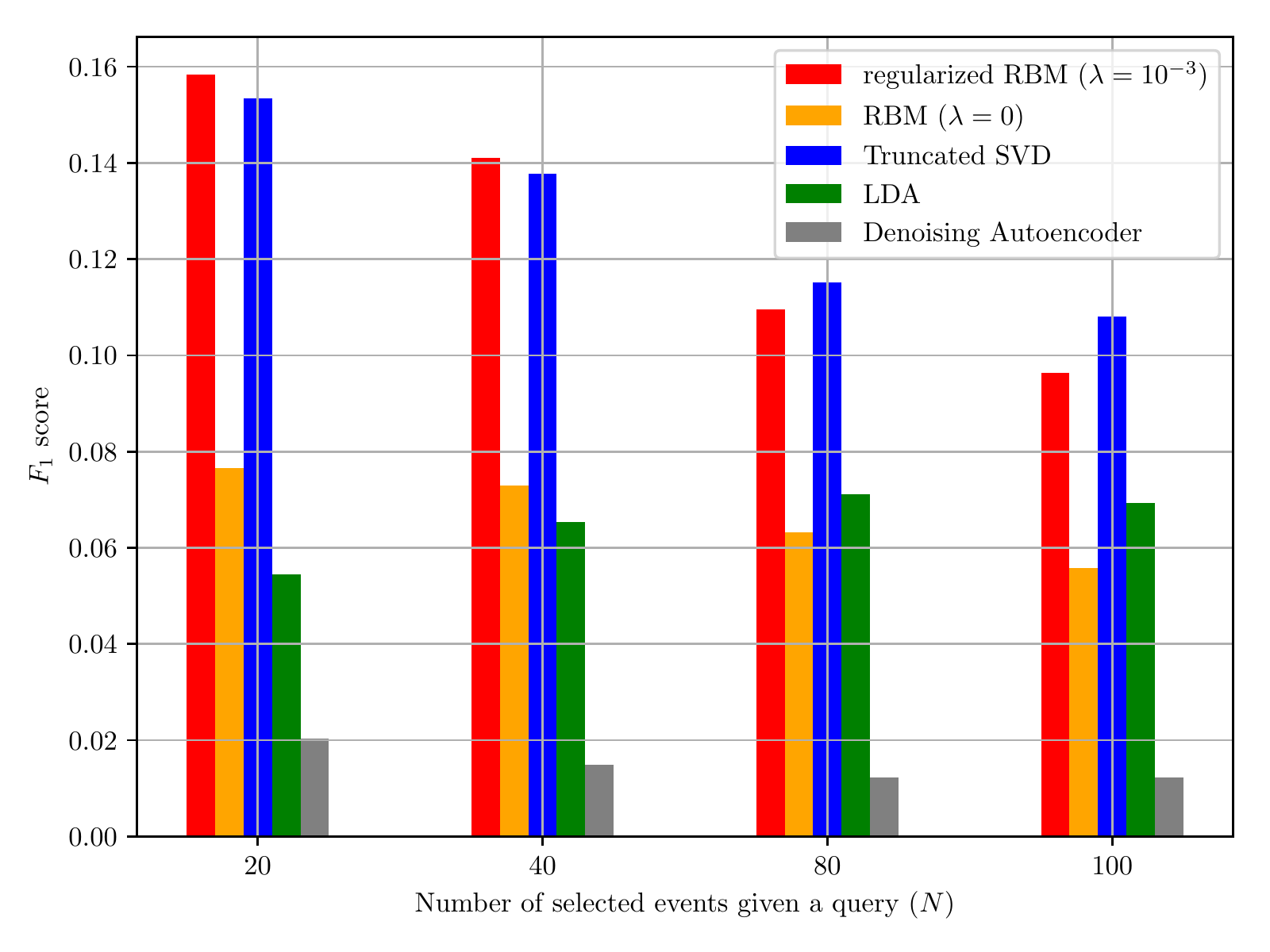}
\caption{$F_1$ scores}
\label{fig:fscore}
\end{subfigure}
\caption{Embeddings of 2,056 crime events projected into a 2D space by t-SNE. (\ref{fig:lam1e-3}), (\ref{fig:lam0}), (\ref{fig:svd}), (\ref{fig:lda}), (\ref{fig:autoencoder}) are generated by regularized RBM ($\lambda = 10^{-3}$), vanilla RBM, truncated SVD, single layer denoising autoencoder and LDA repectively.}
\label{fig:embed}
\vspace{-0.1in}
\end{figure}

In order to intuitively inspect how the high-dimensional embeddings distribute in a 2D space, we first use two-dimensional t-distributed stochastic neighbor embeddings (t-SNE) \cite{tsne.maaten} to project the embeddings with 1,000 binary units into a 2D space. t-SNE is capable of capturing local structure of the high-dimensional data, while also revealing global structures such as the presence of clusters at several scales \cite{tsne.maaten}. Since the embeddings capture events' similarities by vicinity in the Euclidean space, the performance of events embeddings can be visually evaluated by the aggregation degree of labeled events (colorful dots) in such 2D space. Ideally, events that belong to the same crime series should aggregate highly in a local region of the embedding space. In Fig.\ref{fig:embed} we compare \textbf{LDA}, \textbf{truncated SVD}, single layer \textbf{denoising Autoencoder}, vanilla \textbf{RBM}, and \textbf{regularized RBM} ($\lambda = 10^{-3}$) on the crime dataset by observing their distributions visually in t-SNE 2D space. Notice that the each larger and colored dot means individual labeled crime events. \textbf{regularized RBM} outperforms \textbf{LDA}, \textbf{RBM}, \textbf{denoising Autoencoder} with \textbf{truncated SVD} delivering competitive performance. Compare to the vanilla \text{RBM}, the regularization helps improve the quality of the embeddings remarkablely. For instance, the subregion within the red dashes circle over 95\% the green dots, the subregion in Fig.\ref{fig:lam1e-3} is obviously tighter than Fig.\ref{fig:lam0}.

\subsection{Retrieval performance ($F_1$ score)} 

In order to further evaluate the quality of the embeddings accurately, we test them as a retrieval problem, and adopt a basic quality measure, $F_1$ score, widely used in the text mining literature for the purpose of document clustering \cite{fmeasure.steinbach}. The F$_1$ score combines the \textit{Precision} and \textit{Recall} ideas from the information retrieval literature. The precision $P$ and recall $R$ given a query (crime event) with respect to its true class (relevant crime events or crime series) are defined as:
\begin{align*}
&P = \frac{|\text{relevant crime events} \cap \text{retrieved crime events}|}{|\text{retrieved crime events}|}, \\
&R = \frac{|\text{relevant crime events} \cap \text{retrieved crime events}|}{|\text{relevant crime events}|},
\end{align*}
And the $F_1$ score of a query $q$ is defined as $F_1 = 2 \cdot \frac{P \cdot R }{P + R}$. To perform a fair test all the methods, we consider the method with the highest average $F_1$ score to be the best model that maps crime pattern to the embeddings space. 

In the experiment, we first generate embeddings and calculates the pairwise similarities between the crime events according to their cosine distances. And then retrieve the top $N$ keywords given a labeled crime event as a query, as shown in Fig.\ref{fig:fscore}, the RBM with regularization is overwhelmingly better than the others in terms of their $F_1$ score.

\subsection{Keywords selection} 

As we mentioned in the previous section, we cannot obtain the determinate selected features directly from the model. But we do have some indirect approaches to evaluate the quality of the selected variables intuitively. As shown in Fig.\ref{fig:embed_corpus}, Appendix \ref{appen:result}\footnote{See https://arxiv.org/abs/1806.06095 for the appendix.}, we calculate the standard deviations for each of the keywords in the vocabulary over the 2,056 crime events. Most of the keywords in the dataset vary considerably. However, in Fig.\ref{fig:embed_reg}, after reconstructing the same crime events from the RBM with regularization, a large number of keywords have been disabled and are always zero values without variation. 

The remaining activated keywords are very intriguing. We compare top 15 keywords with the highest intensity for both raw dataset and the reconstructed dataset in Fig.\ref{fig:selected_keywords}. Obviously, most of the keywords in Fig.\ref{fig:embed_corpus} are irrelevant to the crime behaviors or even meaningless, but the keywords in Fig.\ref{fig:embed_reg} is the opposite, some of them are even highly consistent with the results we present in Fig.\ref{fig:ngram}, Appendix \ref{appen:motivation}\footnote{See https://arxiv.org/abs/1806.06095 for the appendix.}. The selected words such as \textit{"toyota corolla"}, \textit{"drivers door"}, \textit{"black leather"}, \textit{"silver vehicle"},  \textit{"one ounce"}, \textit{"outside apartment"} or \textit{"hotel"} are the strong indicators to the crime patterns from the perspective of police understanding.

\section{Conclusion}
\label{sec:majhead}

We have presented a novel approach for learning embeddings for crime events with unsupervised feature selection. By imposing a well-designed $\ell_1$ penalty on observed variables' activation probabilities that leads to simple gradient descent based algorithm, our regularized RBMs are able to produce high-quality embeddings as well as eliminate irrelevant and noisy features in observed variables. Additionally, regularized RBMs can select key features without supervision. The selected features are not only highly sparse but also interpretable to human. The techniques introduced in this paper can be also used for learning some other high-dimensional dataset with complex interdependencies between their features. Using real-data, we show promising results on a large-scale real crime dataset comparing to conventional methods for text embedding. 

\newpage
\bibliographystyle{IEEEbib}
\bibliography{refs}

\begin{thebibliography}{10}

\bibitem{triovecevent.zhang}
Chao Zhang,
\newblock ``Triovecevent: Embedding-based online local event detection in
  geo-tagged tweet streams,''
\newblock {\em Proceedings of the 23rd ACM SIGKDD International Conference on
  Knowledge Discovery and Data Mining}, 2017.

\bibitem{event2vec.hong}
Shenda Hong,
\newblock ``Event2vec: Learning representations of events on temporal
  sequences,''
\newblock {\em Asia-Pacific Web (APWeb) and Web-Age Information Management
  (WAIM) Joint Conference on Web and Big Data}, 2017.

\bibitem{word2vec.mikolov}
Tomas Mikolov, Ilya Sutskever, Kai Chen, Greg Corrado, and Jeffrey Dean,
\newblock ``Distributed representations of words and phrases and their
  compositionality,''
\newblock {\em Advances in neural information processing systems}, 2013.

\bibitem{Du2016}
Nan Du and Manuel Gomez-rodriguez,
\newblock ``{Recurrent Marked Temporal Point Processes : Embedding Event
  History to Vector},''
\newblock {\em Kdd}, pp. 1555--1564, 2016.

\bibitem{wang2015}
Tong Wang, Cynthia Rudin, Daniel Wagner, and Rich Sevieri,
\newblock ``{Finding Patterns with a Rotten Core: Data Mining for Crime Series
  with Cores},''
\newblock {\em Big Data}, vol. 3, no. 1, pp. 3--21, 2015.

\bibitem{nlp.stanford}
{Stanford NLP Group},
\newblock ``Tf-idf weighting,''
  https://nlp.stanford.edu/IR-book/html/htmledition/tf-idf-weighting-1.html.

\bibitem{skipgram.mikolov}
Tomas Mikolov, Kai Chen, Greg Corrado, and Jeffrey Dean,
\newblock ``Efficient estimation of word representations in vector space,''
\newblock {\em ICLR Workshop}, 2013.

\bibitem{item2vec.barkan}
Oren Barkan and Noam Koenigstein,
\newblock ``Item2vec: Neural item embedding for collaborative filtering.,''
\newblock {\em IEEE International Workshop on Machine Learning for Signal
  Processing}, 2016.

\bibitem{factorembedding.liang}
Dawen Liang, Jaan Altosaar, Laurent Charlin, and David~M. Blei,
\newblock ``Factorization meets the item embedding: Regularizing matrix
  factorization with item co-occurrence,''
\newblock {\em Proceedings of the 10th ACM conference on recommender systems},
  2016.

\bibitem{context.liu}
Li-Ping Liu,
\newblock ``Context selection for embedding models,''
\newblock {\em Advances in Neural Information Processing Systems}, 2017.

\bibitem{crime.zhu}
Shixiang Zhu and Yao Xie,
\newblock ``Crime incidents embedding using restricted boltzmann machines,''
\newblock {\em 2018 IEEE International Conference on Acoustics, Speech and
  Signal Processing}, 2018.

\bibitem{Luo2010}
Heng Luo, Ruimin Shen, and Cahngyong Niu,
\newblock ``{Sparse Group Restricted Boltzmann Machines},''
\newblock pp. 1--9, 2010.

\bibitem{Halkias2013}
Xanadu Halkias, Sebastien Paris, and Herve Glotin,
\newblock ``{Sparse Penalty in Deep Belief Networks: Using the Mixed Norm
  Constraint},''
\newblock pp. 1--8, 2013.

\bibitem{Ranzato2007}
Marc~Aurelio Ranzato, Christopher Poultney, Sumit Chopra, and Yann Lecun,
\newblock ``{Efficient Learning of Sparse Representations with an Energy-Based
  Model},''
\newblock {\em Advances In Neural Information Processing Systems}, vol. 19, pp.
  1137--1134, 2007.

\bibitem{Ranzato2008}
Marc~Aurelio Ranzato, Y-Lan Boureau, and Yann Lecun,
\newblock ``{Sparse Feature Learning for Deep Belief Networks},''
\newblock {\em Advances in neural information processing systems (NIPS)}, , no.
  Mcmc, pp. 1185--1192, 2008.

\bibitem{hinton2002}
Geoffrey~E. Hinton,
\newblock ``Training products of experts by minimizing constractive
  divergence,''
\newblock {\em Neural Computation}, 2002.

\bibitem{tsne.maaten}
Laurens Van~Der Maaten and Geoffrey Hinton,
\newblock ``{Visualizing Data using t-SNE},''
\newblock {\em Journal of Machine Learning Research}, vol. 9, pp. 2579--2605,
  2008.

\bibitem{fmeasure.steinbach}
Steinbach Michael, Karypis George, and Kumar Vipin,
\newblock ``{A comparison of document clustering techniques},''
\newblock {\em KDD-2000 Workshop TextMining}, 2010.

\bibitem{Fischer2012}
Asja Fischer and Christian Igel,
\newblock ``{An Introduction to Restricted Boltzmann Machines},''
\newblock {\em Lecture Notes in Computer Science: Progress in Pattern
  Recognition, Image Analysis, Computer Vision, and Applications}, vol. 7441,
  pp. 14--36, 2012.

\bibitem{Welling2005}
Max Welling, Michal Rosen-zvi, and Geoffrey~E Hinton,
\newblock ``{Exponential Family Harmoniums with an Application to Information
  Retrieval},''
\newblock {\em Advances in Neural Information Processing Systems (NIPS)}, vol.
  17, pp. 1481--1488, 2005.

\end{thebibliography}

\newpage
\appendix

\section{Motivating application}
\label{appen:motivation}

A key and one of the most challenging tasks in crime analysis is to find {\it related crime events} \cite{wang2015}, which are committed by the same individual suspect or criminal gang. Such series of crimes usually share similar modus operandi (M.O.), for instance, some criminals always break into houses in the late afternoon from backdoor to steal jewels. Finding crime series based on M.O. critically depends on extracting informative features for crime events \cite{wang2015}, which is usually done by the police detectives, however, this is not scalable to larger and ever-growing crime dataset. For instance, in the City of Atlanta, from the year 2013 to 2017, there are a total of 1,096,961 cases with over 800 categories. Besides time, location, and crime category, the most important information in crime events is free text narratives entered by police officers. But there has yet been a tool to automatically extract useful features and information from the free-text narratives in crime events, since the crime narratives are very noisy, unstructured, and even unbalanced written by different police officers, and are sometimes incomplete English sentences since they are written in a haste. That makes existing conventional NLP models such as \textit{"Bag-of-Words"} far from sufficient for this particular task.

\begin{figure}[h!]
\centering
\begin{subfigure}[h]{.32\linewidth}
    \includegraphics[width=\linewidth]{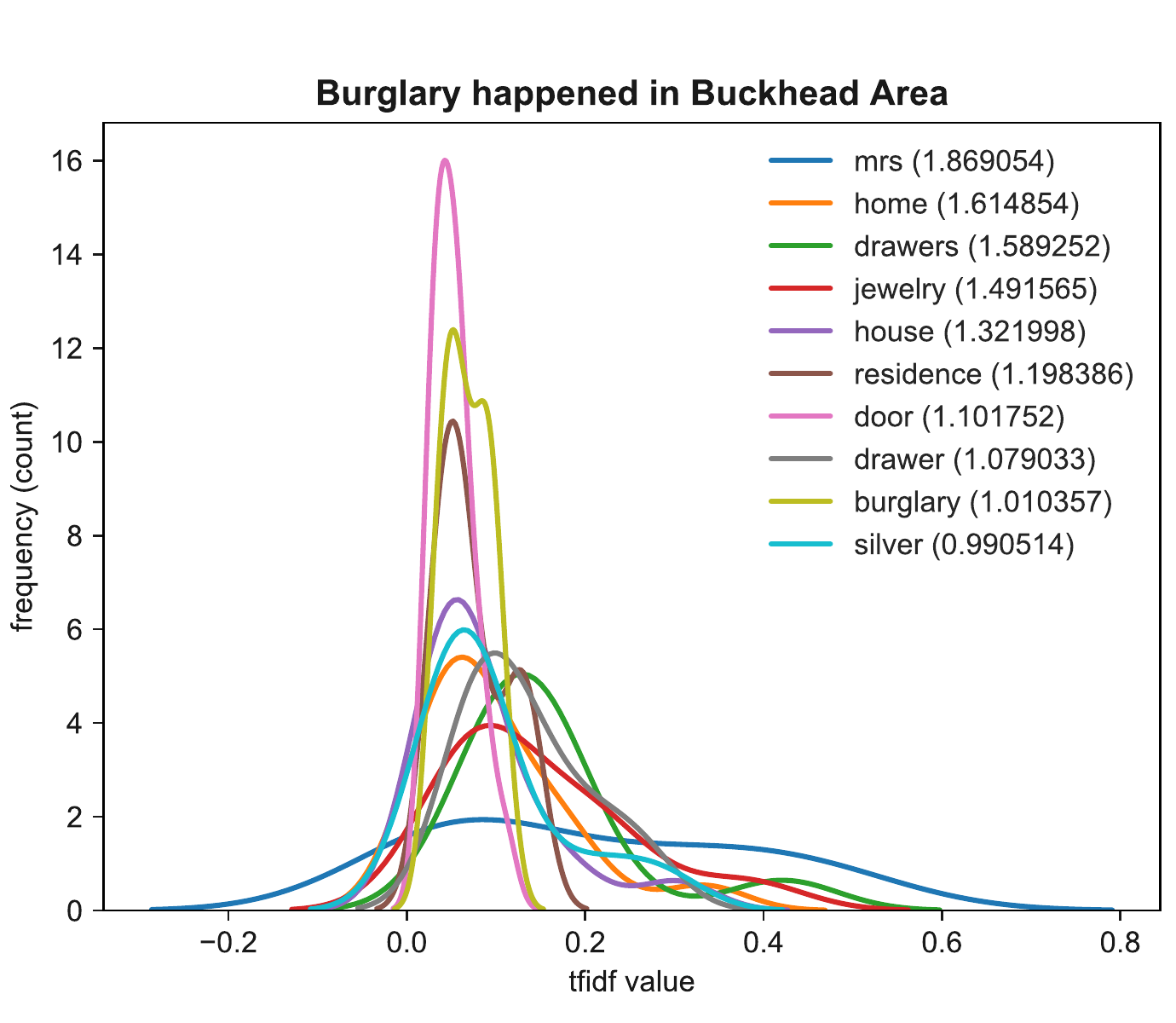}
    \caption{}\label{fig:ngram1}
\end{subfigure}
    \hfill
\begin{subfigure}[h]{.32\linewidth}
    \includegraphics[width=\linewidth]{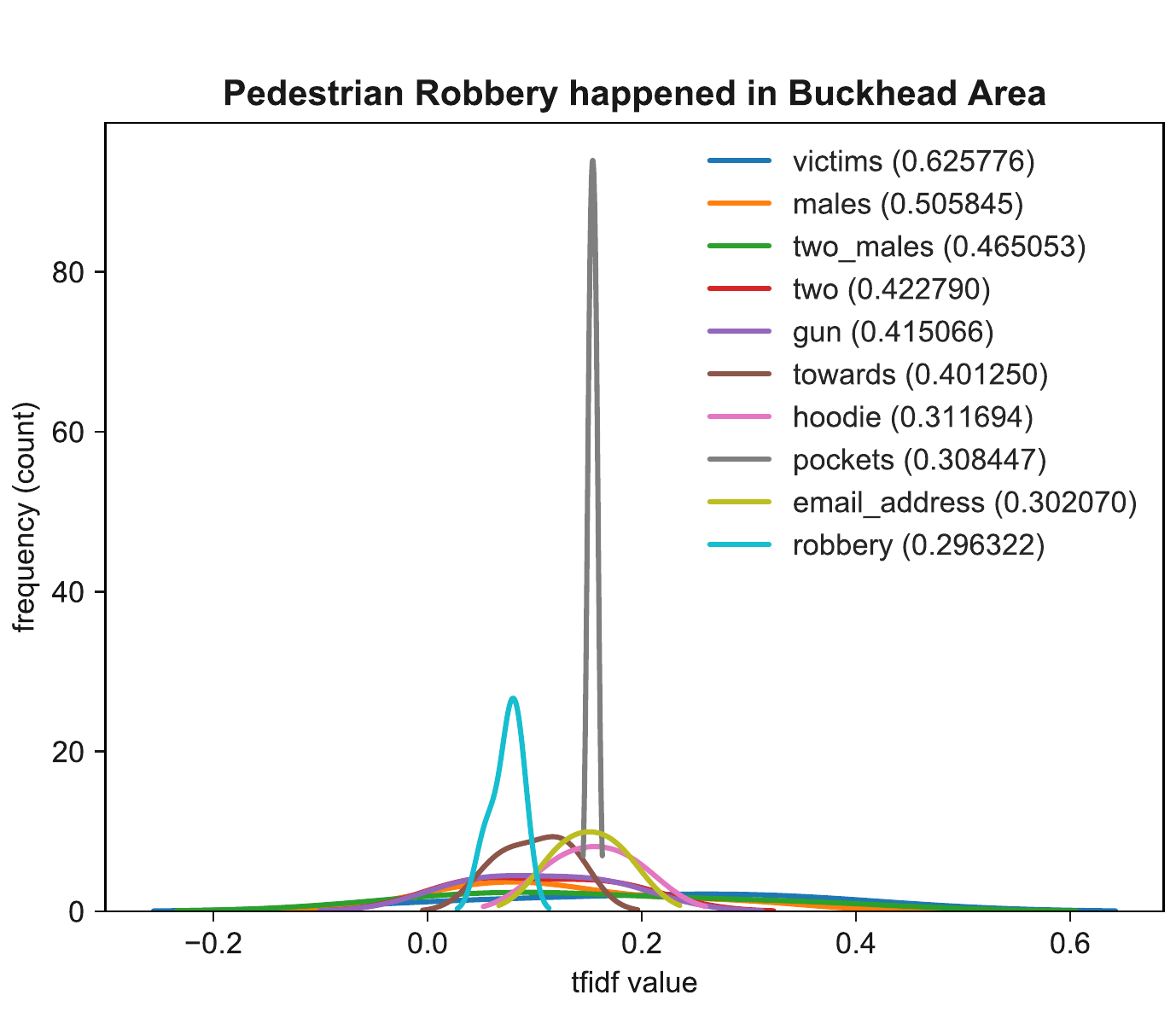}
    \caption{}\label{fig:ngram2}
\end{subfigure}
   \hfill
\begin{subfigure}[h]{.32\linewidth}
    \includegraphics[width=\linewidth]{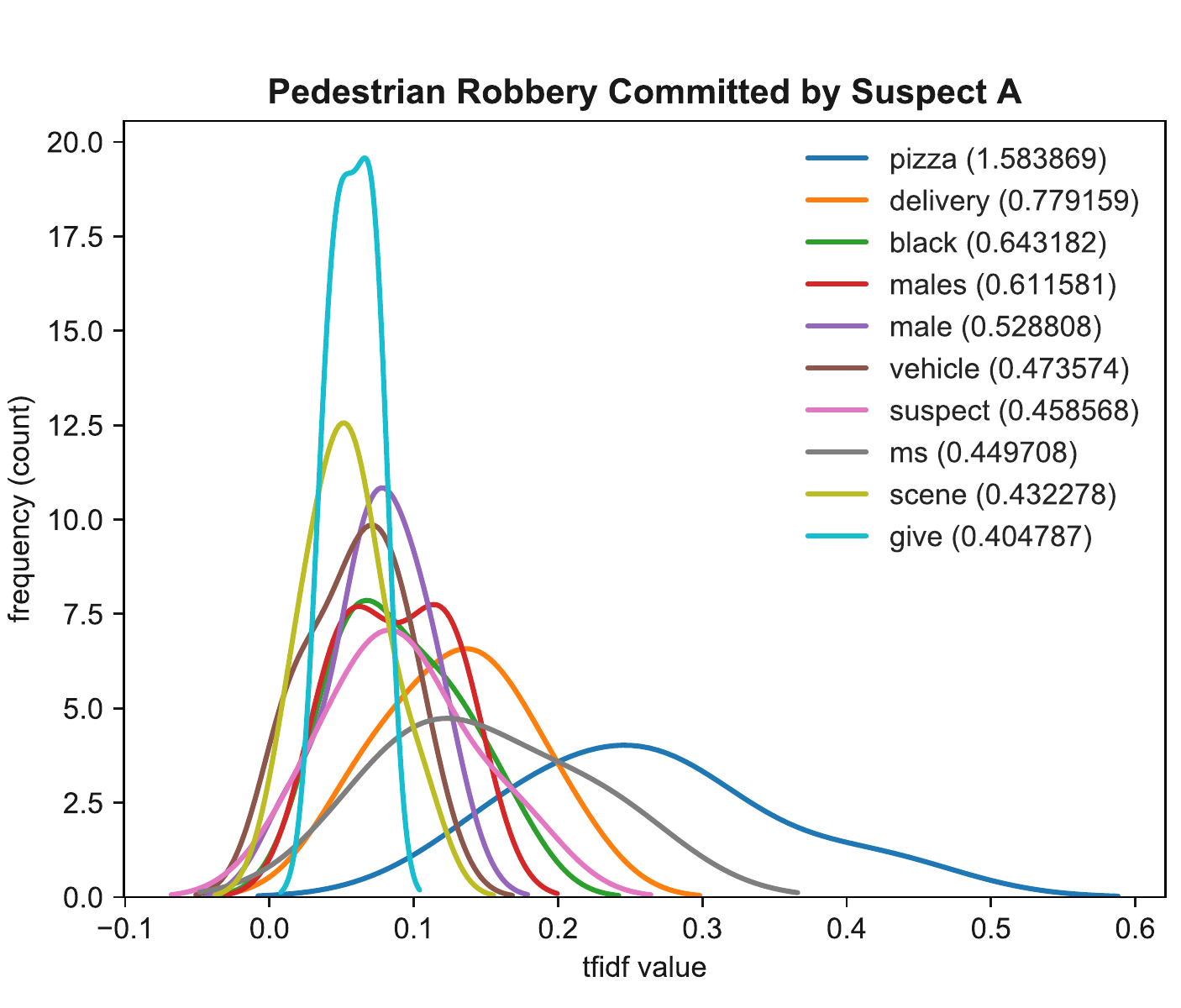}
    \caption{}\label{fig:ngram3}
\end{subfigure}
\vfill
\begin{subfigure}[h]{.32\linewidth}
    \includegraphics[width=\linewidth]{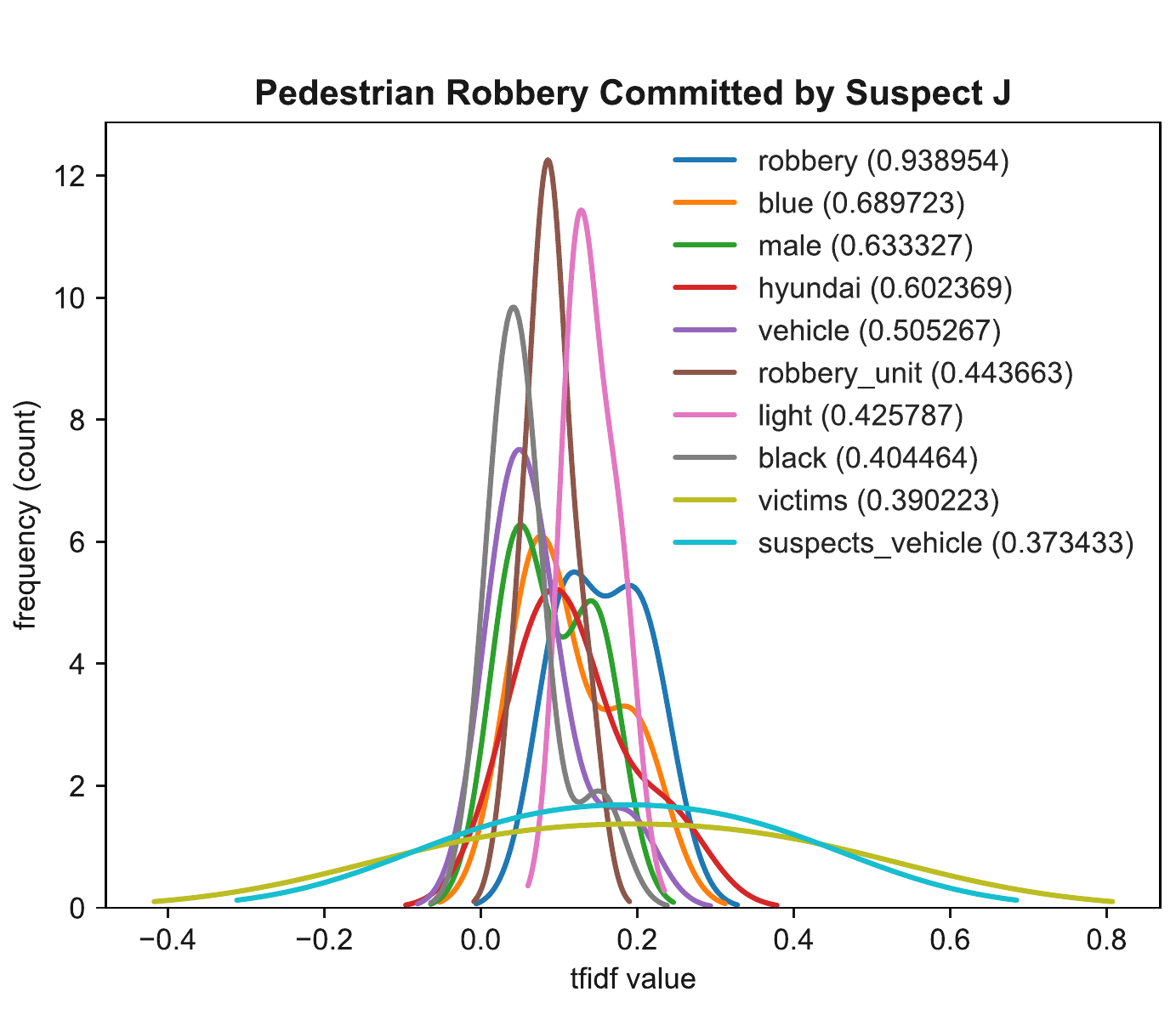}
    \caption{}\label{fig:ngram4}
\end{subfigure}
    \hfill
\begin{subfigure}[h]{.32\linewidth}
    \includegraphics[width=\linewidth]{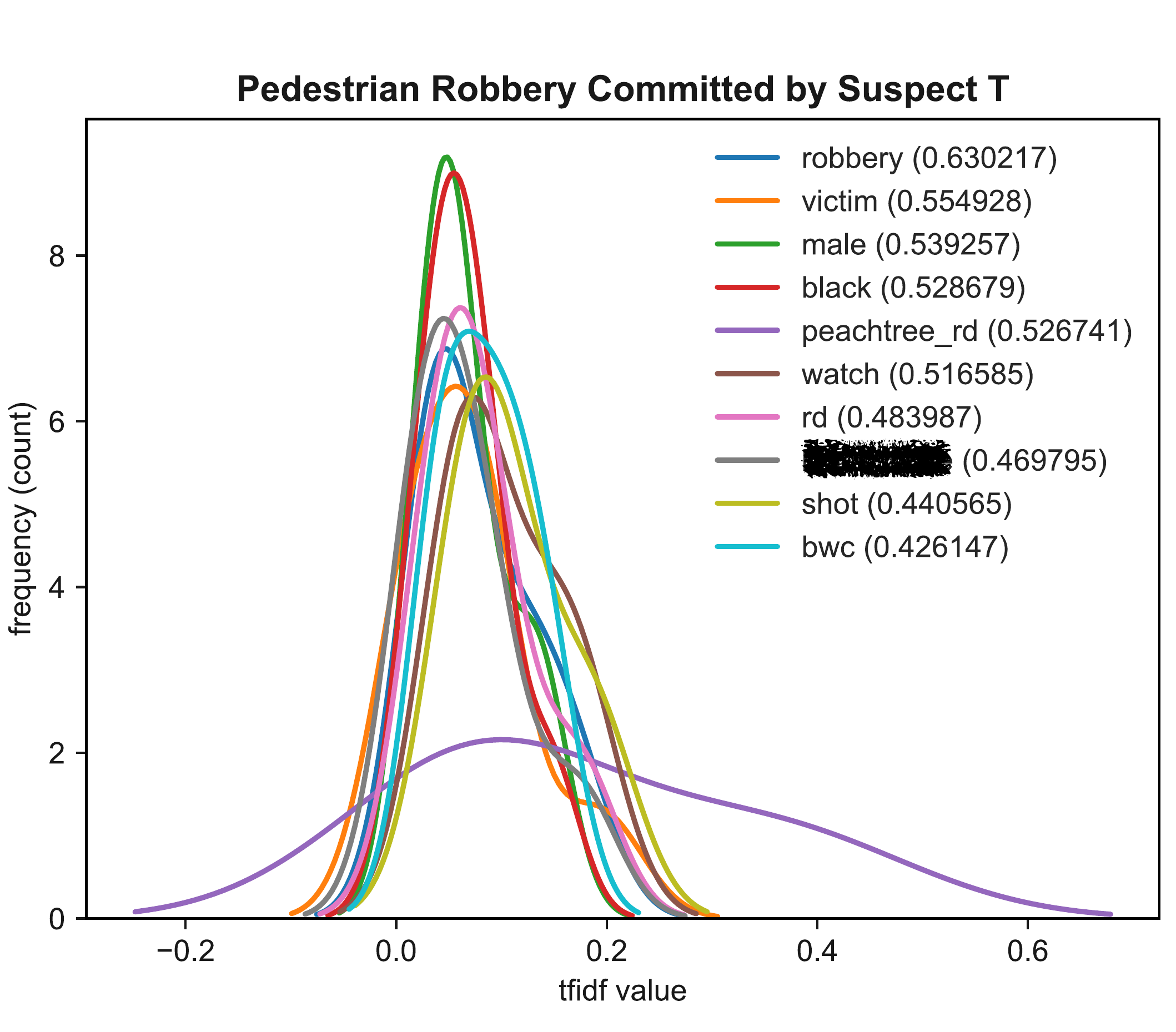}
    \caption{}\label{fig:ngram5}
\end{subfigure}
   \hfill
\begin{subfigure}[h]{.32\linewidth}
    \includegraphics[width=\linewidth]{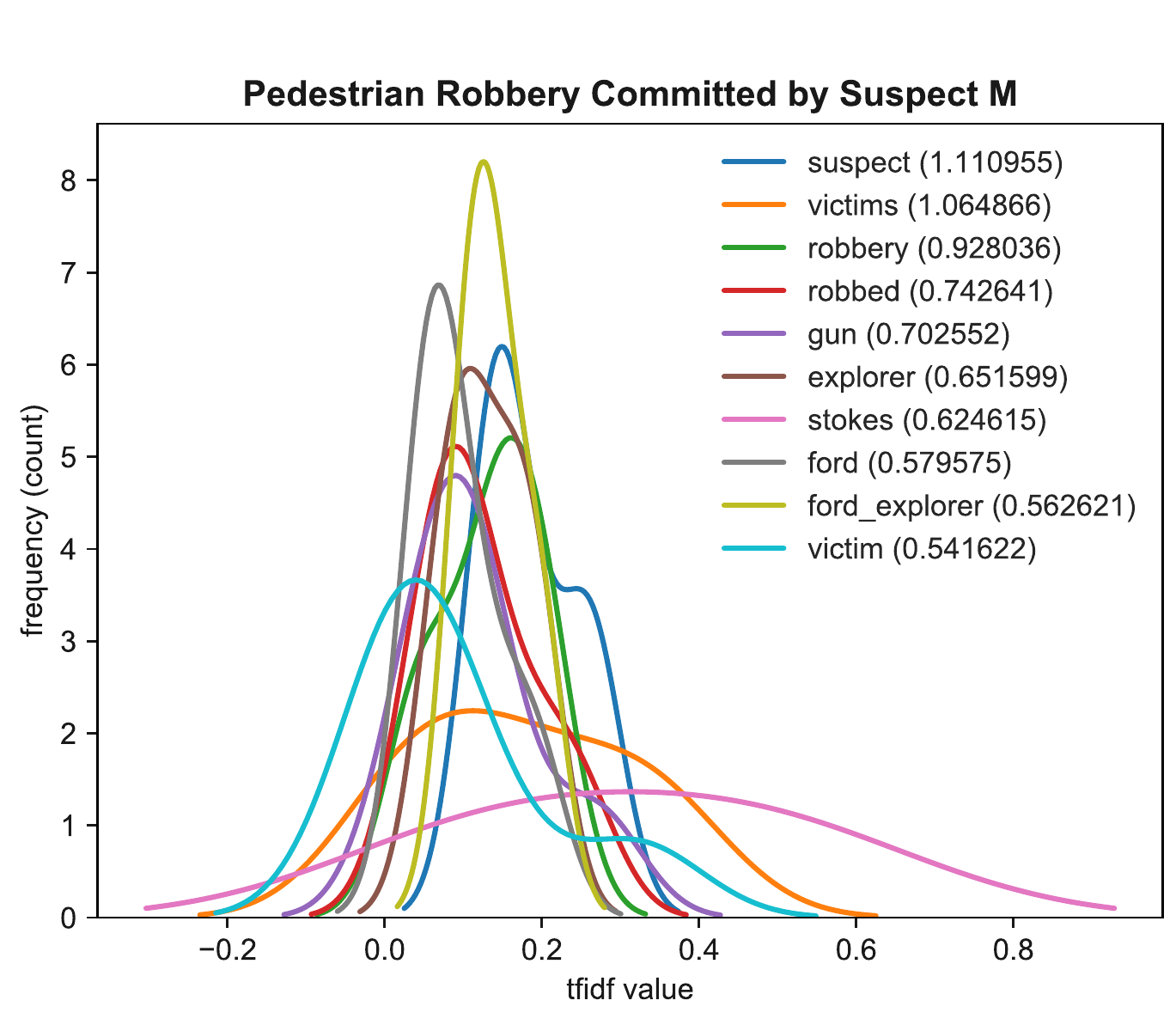}
    \caption{}\label{fig:ngram6}
\end{subfigure}
\caption{Histograms of top 10 high-frequency keywords for 6 crime series. As shown above, the co-occurrence of high-frequency keywords of different crime series have very distinctive pattern. In most cases the specific co-occurrence of keywords reveals key clues of the pattern of a series of crime events, which are able to uniquely identify a crime series.} 
\label{fig:ngram}
\vspace{-0.1in}
\end{figure}

Feature selection is important for embeddings of crime events, because we found that similar M.O. will lead to common keywords to co-occur in the free-text of police reports. As a  strong piece of empirical evidence, we calculate the histogram of tf-idf values \cite{nlp.stanford} for each keyword in each crime series. The results, as shown Fig.\ref{fig:ngram}, somewhat surprising, reveals that different crime series have completely different keywords distribution. And those high-frequency keywords are highly interpretable in terms of the pattern of the crime series as confirmed by the police officers.

\section{Restricted Boltzmann Machines}
\label{appen:rbm}

A Restricted Boltzmann Machine is a two layers neural networks and it can be viewed as a probabilistic graphical model \cite{Fischer2012}. The weights of the network, represented as a matrix $\mathbf{w} = (w_{ij})$, visible bias $\mathbf{b} = (b_i)$ and hidden bias $\mathbf{c} = (c_j)$, which associate $H$ hidden variables $\mathbf{h} = h_{1 \dots H}$ and $V$ observed (visible) variables $\mathbf{x} = x_{1 \dots V}$. Here, for the convenience of demonstrating our application, we assume the real observed variables $\mathbf{x} \in \mathbb{R}^V$ take the Bag-of-Words as input, and binary hidden variables $\mathbf{h} \in \{0, 1\}^H$ produce the embeddings. The models can be easily generalized to other types of variables with different activation probabilities \cite{Welling2005}. A probability is associated with configuration $(\mathbf{x}, \mathbf{h})$ as follows:
\[
p(\mathbf{x}, \mathbf{h}) = \frac{1}{Z} e^{-E(\mathbf{x}, \mathbf{h})},
\]
where  the partition function $Z$ is defined as
$
Z = \sum_{\mathbf{x}, \mathbf{h}} e^{-E(\mathbf{x}, \mathbf{h})},
$
and the energy function $E(\mathbf{x}, \mathbf{h})$ is defined as
\[
E(\mathbf{x}, \mathbf{h}) = -\sum_{i=1}^{V} \sum_{j=1}^{H} w_{ij} h_j \frac{x_i}{\sigma} - \sum_{i=1}^{V} \frac{x_i - b_i}{2 \sigma^2} - \sum_{j=1}^{H} h_j c_j.
\]
Here $\sigma$ is a constant  for controlling the shape of the Gaussian distribution; the RBMs' model parameters, $\theta = (\mathbf{w}, \mathbf{b}, \mathbf{c})$, can be learned by maximizing the log likelihood of marginal probability of a set of observed data, 
\begin{equation}
\begin{aligned}
\log \mathcal{L} (\theta|\{\mathbf{x}^{(k)}\}) = \sum_{k=1}^{K} \log p(\mathbf{x}^{(k)}),
\end{aligned}
\label{eq:log_likelihood}
\end{equation}
where the marginal probability $p(\mathbf{x})$ can be derived as follows:
$
p(\mathbf{x}) = \sum_{\mathbf{h}} p(\mathbf{x}, \mathbf{h}). 
$
There are extensive work on how to fit RBM models and how to perform training using gradient descent. A popular approach is the so-called $k$-step contrastive divergence (CD$_k$) \cite{hinton2002}, which approximates the gradients of the parameters as follow:
\begin{align}
\frac{\partial \log \mathcal{L}(\theta|\mathbf{x})}{\partial w_{ij}}  &= \left < x_i h_j\right >_{p(\mathbf{h}|\mathbf{x})} - \left < x_i h_j\right >_{p(\mathbf{x}, \mathbf{h})},  \label{eq:grad.w} \\
\frac{\partial \log \mathcal{L}(\theta|\mathbf{x})}{\partial b_{i}} &= x_i - \left < x_i \right >_{p(\mathbf{x})} \label{eq:grad.b}, \\
\frac{\partial \log \mathcal{L}(\theta|\mathbf{x})}{\partial c_{j}} &= p(h_j = 1| \mathbf{x}) - \left < p(h_j = 1 | \mathbf{x}) \right >_{p(\mathbf{x})} \label{eq:grad.c}.
\end{align}
where $\left < \cdot \right >_P$ denote the expectation with respect to a distribution $P$.
A special feature of RBM is that the observed and hidden variables conditioned on each other are mutually independent. The conditional distributions can be written as
\[
p(\mathbf{v}|\mathbf{h}) = \prod_{i=1}^{V} p(v_i|\mathbf{h}), \quad
p(\mathbf{h}|\mathbf{v}) = \prod_{j=1}^{H} p(h_j|\mathbf{v}).
\]
And the individual activation probabilities of the hidden and the observed variables are given by
\begin{align}
p(h_j = 1|\mathbf{x}) & = \mbox{sigm} (c_j + \sum_{i=1}^{V} w_{ij} \frac{x_i}{\sigma}), \label{eq:condp.h}\\
p(x_i = x|\mathbf{h}) & = \frac{1}{\sigma \sqrt{2 \pi}} \cdot e^{-\frac{1}{2\sigma^2}(x - b_i - \sigma \sum_{j=1}^{H} w_{ij} h_j)^2} \label{eq:condp.x},
\end{align}
where $\mbox{sigm}$ is a sigmoid function defined as $\mbox{sigm}(x) = 1/(1+e^{-x})$.

\section{Result of keywords selection}
\label{appen:result}

\begin{figure}[h!]
\centering
\vspace{-0.1in}
\begin{subfigure}[h]{1\linewidth}
\includegraphics[width=\linewidth]{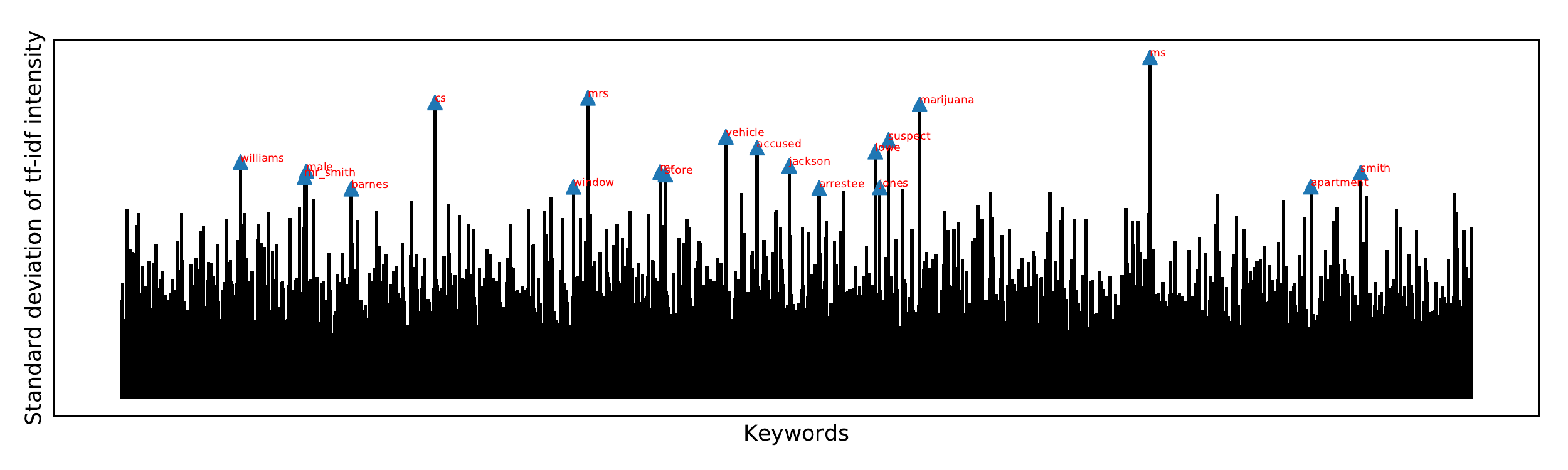}
\caption{RBM ($\lambda=0$)}
\label{fig:embed_corpus}
\end{subfigure}
\vfill
\begin{subfigure}[h]{1\linewidth}
\includegraphics[width=\linewidth]{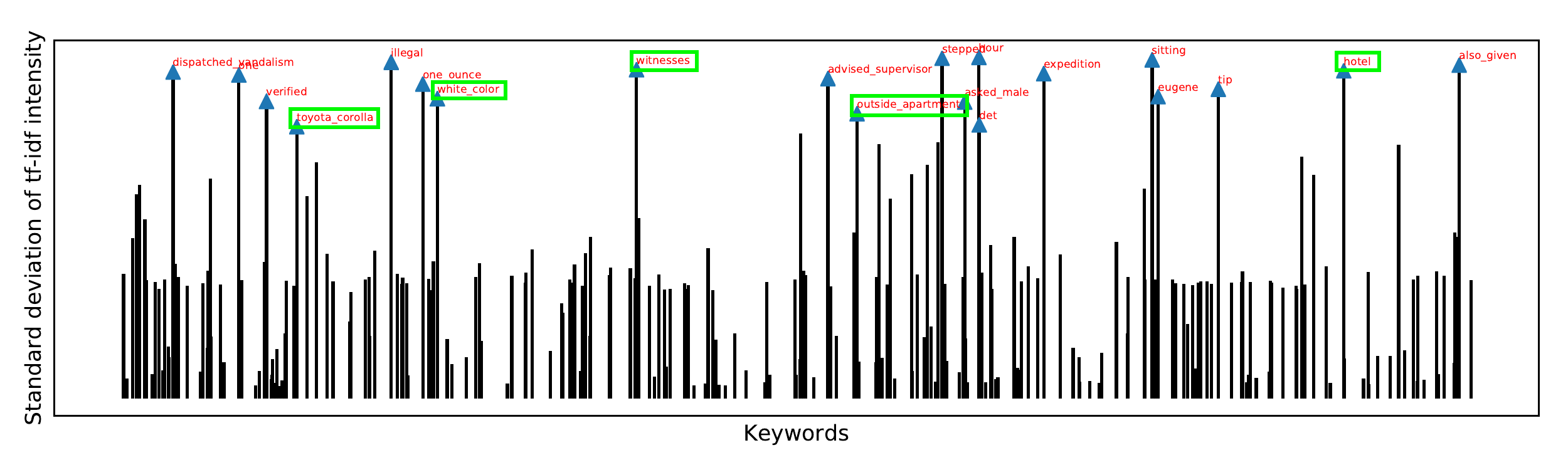}
\caption{regularized RBM ($\lambda=10^{-3}$)}
\label{fig:embed_reg}
\end{subfigure}
\caption{Selected features: (\ref{fig:embed_corpus}): the standard deviations of tf-idf intensity over 2,056 crime events; (\ref{fig:embed_reg}): the same plot as (\ref{fig:embed_corpus}) but the tf-idf intensity is reconstructed by a fitted RBM with $\lambda=10^{-3}$ by taking the raw data as input. Top 15 keywords with the highest standard deviations have been annotated by the side of corresponding bars. The $x$-axis is the 7,038 keywords, and the $y$-axis is the standard deviations of each keyword.}
\label{fig:selected_keywords}
\vspace{-0.1in}
\end{figure}

\end{document}